\documentclass{article}
\usepackage{spconf,amsmath,graphicx}
\usepackage{hyperref}
\usepackage{booktabs,threeparttable}
\usepackage{standalone}
\usepackage{soul}
\usepackage{multirow}
\usepackage{amssymb}
\usepackage{romannum}
\usepackage{xcolor}
\usepackage{color}
\usepackage{bbm}
\usepackage{capt-of}
\usepackage{graphicx,caption}
\usepackage[hang,flushmargin]{footmisc}

\newcommand\blfootnote[1]{%
  \begingroup
  \renewcommand\thefootnote{}\footnote{\noindent #1}%
  \addtocounter{footnote}{-1}%
  \endgroup
}

\title{
Non-Autoregressive Predictive Coding for Learning \\Speech Representations from Local Dependencies
}
\name{Alexander H. Liu \quad Yu-An Chung \quad James Glass}
\address{Computer Science and Artificial Intelligence Laboratory\\
Massachusetts Institute of Technology\\
Cambridge, MA 02139, USA\\
         \small{\texttt{\{alexhliu, andyyuan, glass\}@mit.edu}
         }}
\begin{document}
\ninept
\maketitle
%


\begin{abstract}
Self-supervised speech representations have been shown to be effective in a variety of speech applications.
However, existing representation learning methods generally rely on the autoregressive model and/or observed global dependencies while generating the representation.
In this work, we propose Non-Autoregressive Predictive Coding (NPC), a self-supervised method, to learn a speech representation in a non-autoregressive manner by relying only on local dependencies of speech.
NPC  has a conceptually simple objective and can be implemented easily with the introduced Masked Convolution Blocks.
NPC offers a significant speedup for inference since it is parallelizable in time and has a fixed inference time for each time step regardless of the input sequence length.
We discuss and verify the effectiveness of NPC by theoretically and empirically comparing it with other methods. We show that the NPC representation is comparable to other methods in speech experiments on phonetic and speaker classification while being more efficient.

\end{abstract}

\vspace{-5pt}
\vspace{-3pt}
\section{Introduction}
\label{sec:intro}
\vspace{-3pt}
Speech representation learning aims to extract high-level representations from surface features such as waveforms or spectrograms.
Ideally, these representations make hidden information in speech (such as phonetic content and speaker characteristics) more accessible to downstream tasks.
While speech representations can be defined by different transformations on the surface feature, recent researches~\cite{oord2018representation,riviere2020unsupervised,schneider2019wav2vec,baevski2020vq,baevski2020wav2vec,chung2019unsupervised,pascual2019learning,wang2020unsupervised,liu2020mockingjay} have shown great success by combining neural networks and self-supervised learning (where learning targets can be derived from the input itself).

Contrastive Predictive Coding (CPC)~\cite{oord2018representation} is one such approach whereby the surface feature sequence is first encoded into a latent representation by an encoder network, and an autoregressive model is used to summarize the past latent sequence into a higher-level representation and use it to predict future latent representations.
CPC and its extensions~\cite{riviere2020unsupervised,schneider2019wav2vec,baevski2020vq,baevski2020wav2vec}, have proven to be effective for learning expressive and robust representations of speech.

Instead of targeting future latent representations, Autoregressive Predictive Coding (APC)~\cite{chung2019unsupervised} suggests that simply predicting future surface features is suitable for learning an effective representation of speech.
APC can be improved by enforcing constraints that information from past sequences be stored in the  representation~\cite{chung2020improved} or by imposing an information bottleneck via vector quantization~\cite{chung2020vector}.

Inspired by the left-to-right nature of speech, both CPC and APC achieve self-supervision by using future features in a uni-directional ordered learning.
Masked Language Modeling (MLM)~\cite{devlin2019bert}  relaxes this constraint and uses a different self-supervised learning strategy whereby parts of the input sequence are randomly masked and set to the predicting target, allowing models to input the entire surface feature sequence without seeing the target and derive representation from the context information.
In practice, a bidirectional RNN~\cite{wang2020unsupervised} or Transformer encoder~\cite{liu2020mockingjay} can be employed in learning speech representation through MLM.
\blfootnote{Code available at \footnotesize{\protect\url{https://github.com/Alexander-H-Liu/NPC}}}

To introduce our work, we first formulate our task and mark two properties of the aforementioned methods.
The goal is to derive a high-level representation $(h_1, h_2, ..., h_T)$ from the surface feature sequence of audio  $(x_1,x_2,...,x_T)$ with length $T$.
In APC and CPC, the representation $h_t$ at time $t$ is learned by predicting the unseen future frame $x_{t+n}$ (or its latent) based on the current frame $x_t$ and the previous representation $h_{t-1}$.
These methods are 1) inherently \textit{autoregressive}: the previous representation $h_{t-1}$ is required at each timestep; and 2) incorporating \textit{global dependency}:  $h_{t-1}$ encodes all the past inputs $(x_1,...,x_{t-1})$, making $h_t$ to depend on $(x_1,..,x_t)$.
These properties also apply to MLM\footnote{The autoregressive property of MLM can be eliminated by transformer~\cite{vaswani2017attention} at a cost of increasing time complexity in terms of sequence length.}, but with a stronger global dependency as the full input sequence is always observed, i.e. $h_t$ depends on $(x_1,...,x_T)$ for any $t$.
Note that these two properties have a huge impact on the efficiency of representation models.
To be more specific, the autoregressive property implies that the extraction process cannot be parallelized in time, and relying on global dependency results in time complexity bounded by the input sequence length as we verify later in our experiments (Sec.~\ref{exp:compare}).

To this end, we propose Non-Autoregressive Predictive Coding (NPC) to learn representations in a \textit{non-autoregressive} manner and observing only the \textit{local dependency} of speech.
Without the autoregressive property, NPC offers a significant speedup for deriving speech representation by parallelizing in time.
By observing only local dependencies, NPC allows representations to be derived efficiently regardless of the input sequence length, which is useful for downstream tasks requiring low latency such as streaming speech recognition.
Furthermore, we show that representations derived by NPC, relying only on local dependencies and a non-autoregressive model, is empirically comparable to different prior works.

\begin{figure*}[ht]
\centerline{\includegraphics[width=16cm]{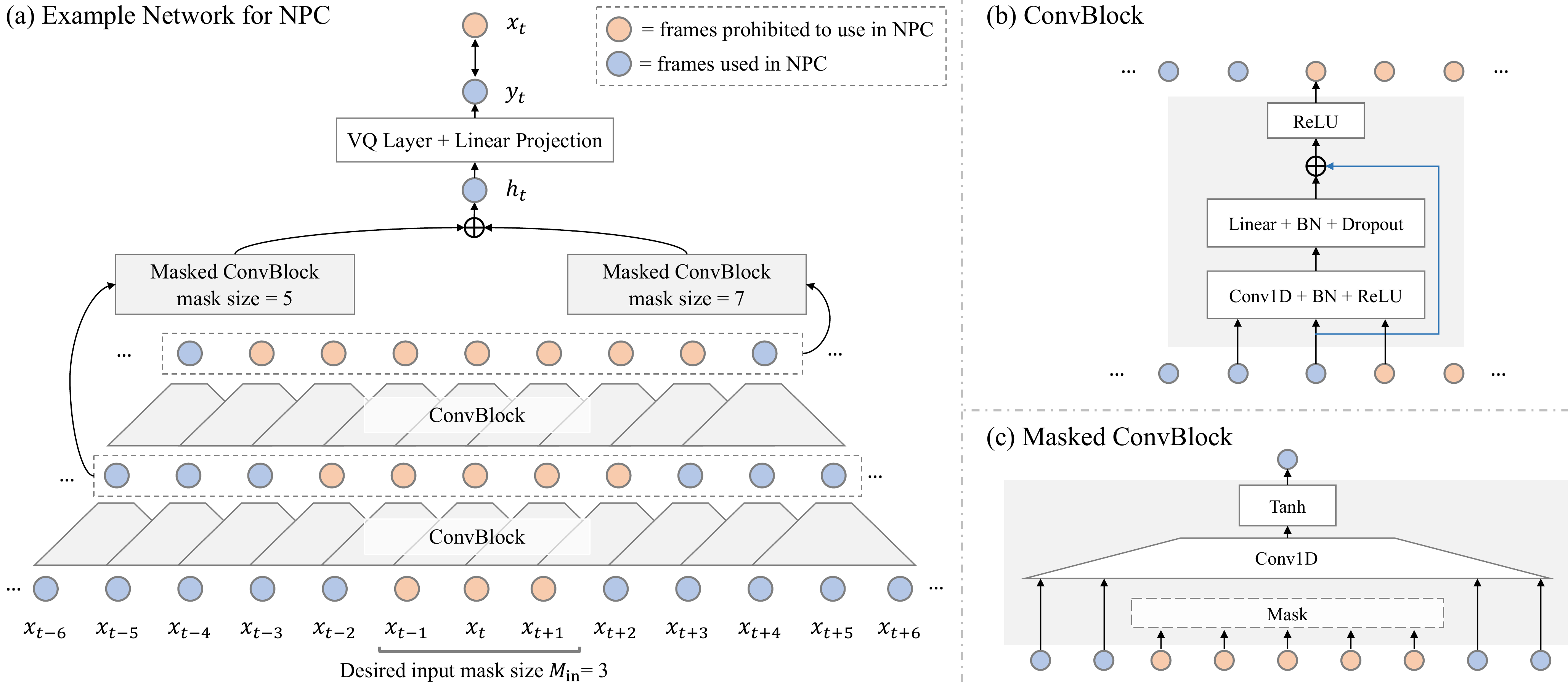}}
\vspace{-2pt}
\caption{\small Illustration of NPC at time $t$ with desired input mask size $M_{\text{in}} = 3$ on an example network having receptive field size $R=13$.
In all figures, orange nodes represent the frames that contain information of the target frame $x_t$, therefore should not be used for prediction; blue nodes are the rest of frames that can be used.
(a) An example of 2-layered feedforward network for NPC.
Input frames are processed by layers of ConvBlock, features from Masked ConvBlock at each layer are summed up to the context representation $h_t$, which will be passed into Vector Quantization layer followed by a linear projection predicting $y_t$ to match the target frame $x_t$.
(b) ConvBlock applies a CNN along the time axis, resulting target-related information to spread to its neighbor at next layer.
(c) Masked ConvBlock generates representation only based on unmasked frames containing no prohibited information.
}
\vspace{-6pt}
\label{fig:overview}
\end{figure*}

\vspace{-3pt}
\section{Proposed method}
\label{sec:method} 
\vspace{-3pt}
\subsection{Non-autoregressive Predictive Coding}
\label{subsec:npc}

To derive the high-level feature $h_t$ at time $t$ without a global dependency or autoregressive property, we restricted it to depend only on the neighbors of $x_t$ in time within the receptive field $(x_{t-r},...,x_{t},...,x_{t+r})$ of size $R = 2r+1$.
While any model architecture with a fixed-size receptive field can be applied for the purpose, we stacked Convolution Blocks (ConvBlock, Fig.~\ref{fig:overview}(b)) to build the representation extraction model  in this work.

To ensure that the high-level feature $h_t$ is indeed representative of $x_t$, it is linearly transformed into $y_t$ to predict $x_t$.
Following previous work~\cite{chung2020vector,baevski2020vq,chorowski2019unsupervised}, we adopt a Vector-Quantization~\cite{van2017neural} layer before the linear projection to serve as an information bottleneck on $h_t$ to yield a better representation.
The objective of NPC is to minimize the \textit{L}1 difference between surface feature $x_t$ and the prediction $y_t$ based on $h_t$ for all time steps
\begin{equation}
    \sum_{t=1}^T \left| y_t - x_t  \right|.
\end{equation}

Note that the target $x_t$ of representation $h_t$ is in the receptive field $(x_{t-r},...,x_{t},...,x_{t+r})$, which might cause $h_t$ to be uninformative as the network can implicitly learn to copy the target directly from the input.
Therefore, NPC requires an additional restriction where the target and its close neighbors in time cannot be observed by $h_t$.
Concretely, given the receptive field $(x_{t-r},...,x_{t},...,x_{t+r})$ of the high-level representation $h_t$, the nearest $2m$ neighbors of $x_t$ and itself, i.e. $(x_{t-m},...,x_t,...,x_{t+m})$, cannot be observed, forming a desired input mask size $M_\text{in} = 2m+1$ for $h_t$.
As the receptive field of each layer in the model varies, the desired mask size changes accordingly, e.g., the choice of ConvBlock with receptive field of size 3 results in the desired mask size to increase by 2 (see orange nodes in Fig.~\ref{fig:overview}(a)(b)).


\subsection{Masked Convolution Blocks for NPC}
\label{subsec:mask_conv}

To implement the desired restriction, we introduce the Masked Convolution Block (Masked ConvBlock), where the kernel-wise convolution operation can be written as
\begin{equation}
    (W \odot D)\ast Z
\end{equation}
with $Z\in \mathbb{R}^{T\times d}$ denoting the intermediate features from model with sequence length $T$ and dimension $d$, $W\in \mathbb{R}^{k\times d}$ denoting the learnable kernel weight with size $k$, and $D \in \{0,1\}^{k\times d}$ denoting the mask with each element $ d_{ij} = \mathbbm{1}_{i\leq \frac{k}{2}-m} + \mathbbm{1}_{i\geq \frac{k}{2}+m}$.
For example, Fig.~\ref{fig:overview}(c) illustrated a Masked ConvBlock with $k=7$ and $m=2$.  

The Masked ConvBlock prevents high-level feature $h_t$ from observing any surface feature within the desired input mask.
Moreover, it can be applied to any intermediate level feature as long as the desired mask size can be calculated at each layer.
In practice, we find this property valuable as it allows aggregation of representations at different depths.


\begin{figure*}[ht]
  \begin{minipage}{.37\textwidth}
    \captionsetup{width=.9\linewidth}
    \centerline{\includegraphics[width=0.95\linewidth]{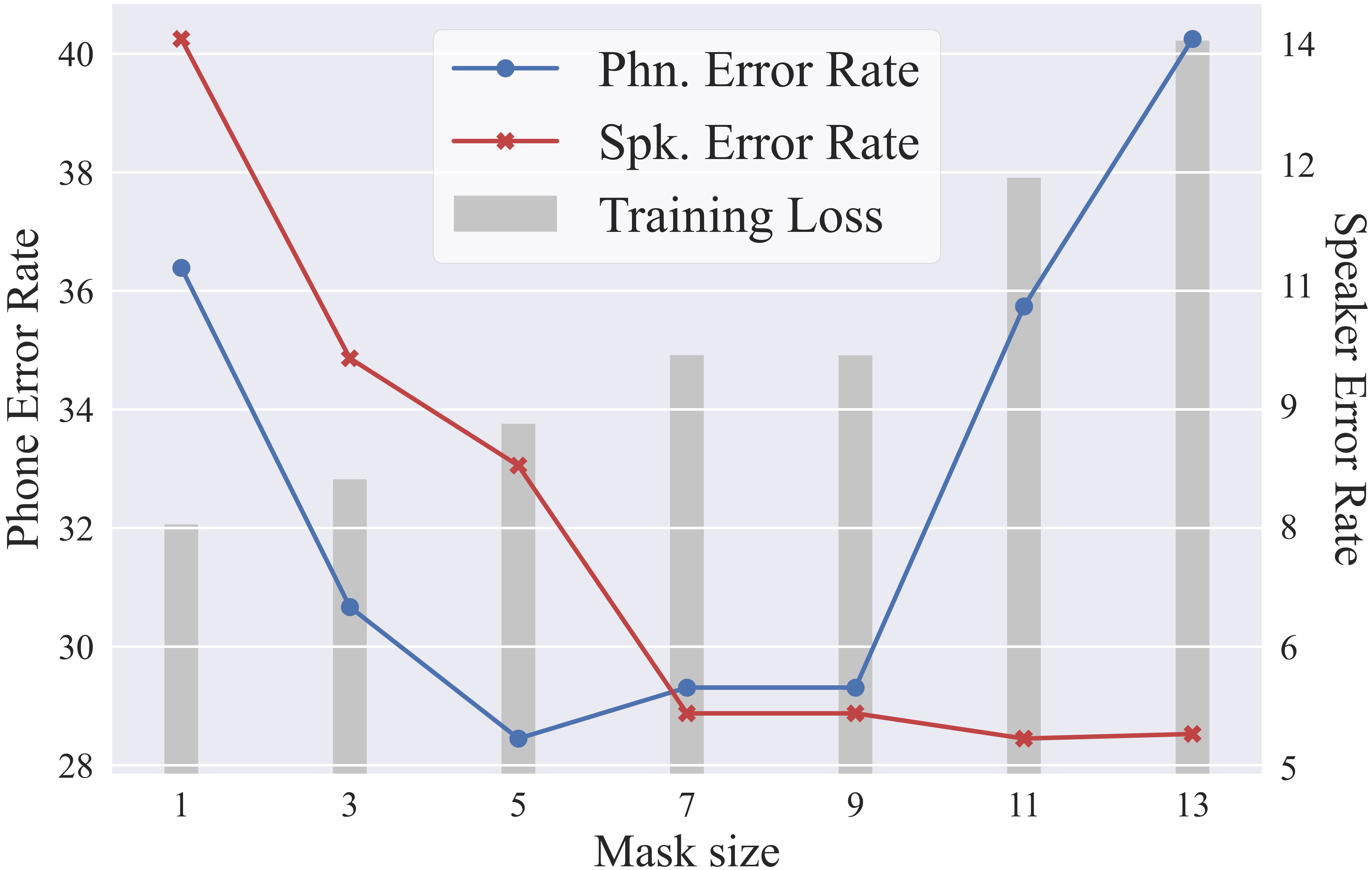}}
    \caption{\small Phone/speaker error rate and training loss with respect to different mask size on 2-layer NPC with receptive field size $R=23$.}
    \label{fig:mask}
  \end{minipage} 
  \begin{minipage}{.37\textwidth}
    \captionsetup{width=.9\linewidth}
    \includegraphics[width=0.93\linewidth]{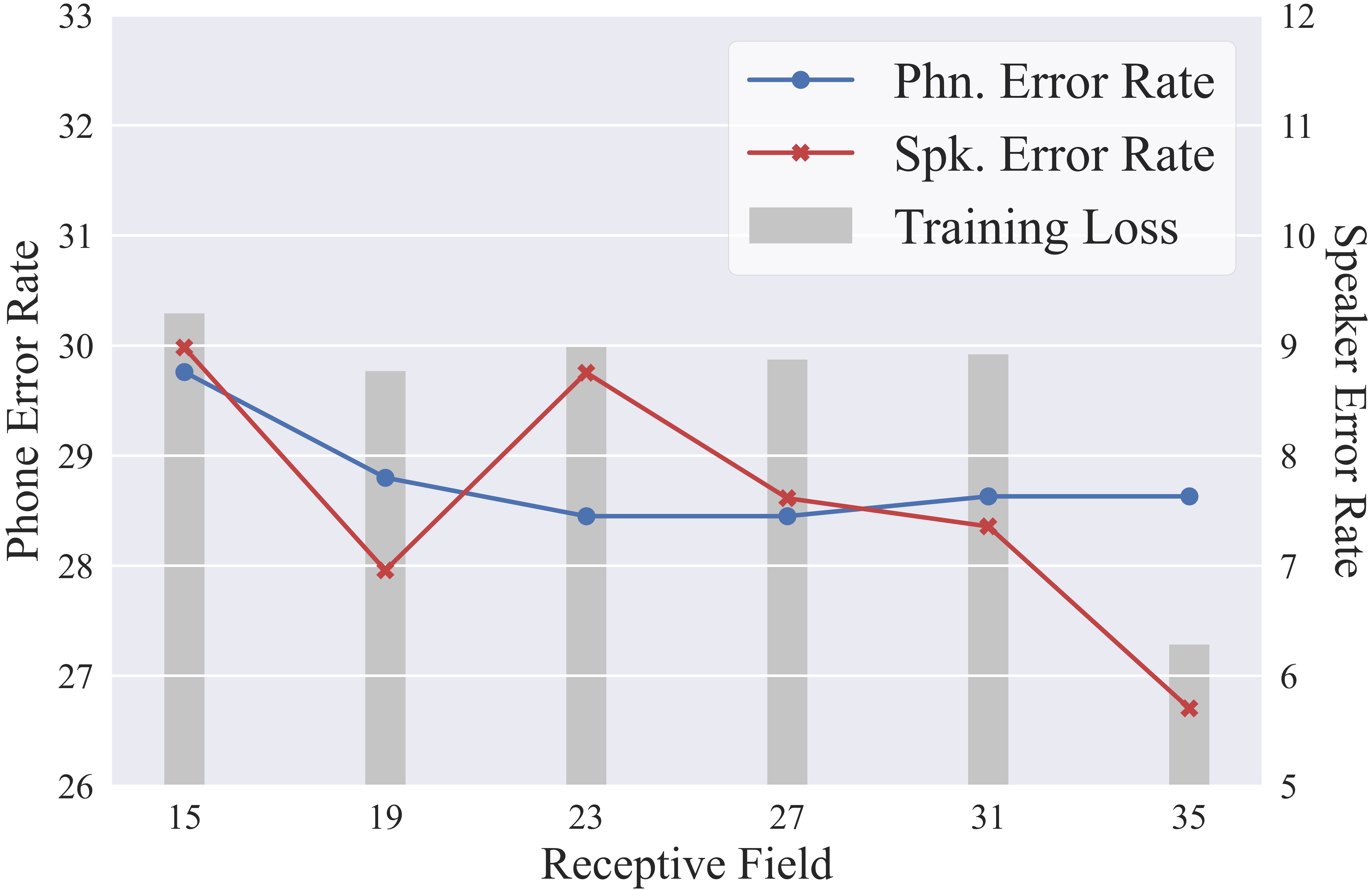}
    \caption{\small Phone/speaker error rate and training loss with respect to different receptive field size on 2-layer NPC with input mask size $M_\text{in}=5$.}
    \label{fig:kernel}
  \end{minipage} 
  \begin{minipage}{.25\textwidth}
    \captionsetup{width=.9\linewidth}
        \small
        \centering
        \begin{tabular}{ l| c}
        \toprule
        \multicolumn{1}{l|}{Method}  & PER  \\ \hline
         NPC 4-layer &  27.2 \\\hline
          - remove 1 layer &  27.7 \\
          - remove 2 layer  & 28.8  \\
          - remove VQ layer & 27.9 \\
          - Single MaskedConv & 29.7 \\
        \bottomrule
        \end{tabular}
        \captionof{table}{\small Ablation study on NPC with input mask size $M_\text{in}=5$, receptive field size $R=27$. Single MaskedConv indicates applying Masked ConvBlock at the last layer only.}
        \label{table:ablation}
  \end{minipage}
\vspace{-10pt}
\end{figure*}

\section{EXPERIMENTS}

\subsection{Setup}

\noindent \textbf{Self-supervised learning.} We learn speech representations from the clean 360-hour subset of LibriSpeech~\cite{panayotov2015librispeech}.
An 80-dimensional log Mel spectrogram is selected as the surface feature of speech.
Unless otherwise specified, each channel is normalized to have zero mean and unit variance across the same utterance.
For the NPC model, we use multi-layer convolution networks, each layer consists of a ConvBlock and Masked ConvBlock as described in Fig.~\ref{fig:overview}.
Given a desired receptive field $R$, since ConvBlocks have a fixed receptive field of 3, the kernel size of Masked ConvBlock can be set to $R-2\times L$ where $L$ is the depth of NPC model.
Throughout our experiments, we fix the dimension of representation and all the intermediate layers to be 512.
We use the Gumbel-softmax vector quantization layer described in ~\cite{baevski2020vq} with a group of 4 codebooks and each group consists of 64 codewords.
We train NPC using Adam~\cite{kingma2015adam} with a learning rate of $10^{-3}$ and a batch size of 32 for 50 epochs.

\noindent \textbf{Evaluation of representation.} We follow the previous works~\cite{oord2018representation,chung2019unsupervised,liu2020mockingjay} to define the ``effectiveness'' of representations as the accessibility to hidden information in speech, i.e., their linear separability with respect to the underlying phonetic labeling and speaker identity.
The model pre-trained with NPC on LibriSpeech is fixed and used to extract representations from the Wall Street Journal corpus (WSJ)~\cite{paul1992design} for the following tasks where the setting aligns with the previous work~\cite{chung2020vector}.
For phone classification, we used 90\% of the utterances in the standard \texttt{si284} split to train a linear classifier, the remaining 10\% as the validation set, and reported the frame-wise test accuracy on \texttt{dev93}.
For speaker classification, the extracted representations are averaged utterance-wise to serve as the input of linear classifier.
We consider the first 259 speakers in \texttt{si284} and used 80\% of utterances as the training set, 10\% as the validation set, and reported the frame-wise test accuracy on the last 10\%.
All reported numbers are averaged over 3 run with negligibly small variance.

\vspace{-6pt}
\subsection{Importance of mask size and receptive field size}

We start with experiments on the choice of hyperparameters for NPC: the desired mask size $M_\text{in}$ and the receptive field size $R$.

\vspace{2pt}
\noindent \textbf{Size of the desired input mask.}
Fig.~\ref{fig:mask} shows the result of varying the mask size with a fixed receptive field of size $23$, i.e. restricting inputs to be 11 frames on both sides of the target.
Intuitively, increasing the mask size will increase the difficulty of predicting the target frame, and the training loss increases accordingly as a consequence.
It can be observed that with the mask size lower than 5, NPC representations begin to lose speaker and phonetic accuracy despite having a lower loss, which verifies our assumption in Sec.~\ref{subsec:npc} where observing the target and its close neighbor will result in a less informative representation.
On the other hand, a dramatic increase in phone error rate but not the speaker error rate is observed as the mask size exceeds 9, indicating that proper constraint on mask size is important for NPC to capture phonetic content.
This matches the fact that phonetic content may change within a short time period while speaker characteristics tend to persist across time, hence are not affected by a larger mask.

\vspace{2pt}
\noindent \textbf{Size of the receptive field.}
Fig.~\ref{fig:kernel} shows the result of varying the receptive field size with a fixed input mask size of $5$, i.e. the representation is learned without observing the target and 2 adjacent frames on both sides.
It is important that the phone error rate and speaker error rate do not differ with respect to the size of receptive field as much as the mask size, indicating that the mask size is the more important hyperparameter introduced by NPC.
Moreover, the fact that phone error rate does not decrease significantly as the receptive field grows verified our claim that local dependency is sufficient for learning speech representations to a certain degree.

\subsection{Importance of model architecture}

To verify the importance of the model architecture, we performed an ablation study and list the results in Table~\ref{table:ablation}.
We note that the difference in speaker error rate is not significant and we only report phone error rate.
We start with a 4-layer NPC model with receptive field size $R=27$ and input mask size $M_\text{in}=5$.
By either reducing the depth of the NPC model or removing the vector quantization layer, the phone error rate slightly increased but varied no more than 1.6\%. 
In contrast, phone error rate drops over 2\% when applying the Masked ConvBlock on the last layer only (29.7). 
Nevertheless, we observe that none of the architectural decisions have a huge impact on NPC as we also saw for the input mask size $M_\text{in}$.
We believe this demonstrates the robustness of the NPC model in terms of architecture.

\renewcommand{\arraystretch}{1.2}

\begin{table*}[ht]
\small
\centering
\begin{threeparttable}
\caption{ Efficiency and performance of different self-supervised methods. All representation have dimension of 512, speaker-wise normalized log Mel spectrogram is used as the surface feature. All numbers in the phone and speaker error rate columns except those of NPC are directly taken from~\cite{chung2020vector}. See Sec.~\ref{exp:compare} for more setup details.}
\begin{tabular}{l | c c | c c | c c }
\toprule
\multirow{2}{*}{Method} & \multirow{2}{*}{Network} & Frame  & Theoretical\textsuperscript{\textdagger}  & Empirical\textsuperscript{\textsection} &  Phone & Speaker\\ 
& & dependency & complexity  & inference time & error rate & error rate\\ \hline \hline
log Mel-spectrogram &  - & -  & -  & - &  50.3 & 17.6 \\ \hline
CPC \cite{oord2018representation} & \multirow{4}{*}{3-layer GRU} & \multirow{4}{*}{Left-to-right}  &  \multirow{4}{*}{$\mathcal{O}(T \cdot d^2)$}  &  \multirow{4}{*}{29x} &  34.1 & 9.7 \\
APC \cite{chung2019unsupervised} &  &   &  &  &  33.3 & 8.5 \\
MT-APC \cite{chung2020improved} &  &   &  &  &  30.5 & 7.3 \\
VQ-APC \cite{chung2020vector} &  &   &  &  &  28.4 & 5.5 \\ \hline
RNN-MLM \cite{wang2020unsupervised} & 3-layer Bi-GRU & \multirow{2}{*}{Global} & $\mathcal{O}(T \cdot d^2)$ &  72x & 32.4 & 6.2 \\
Transformer-MLM \cite{liu2020mockingjay} & 3-layer Transformer &  & $\mathcal{O}(T^2 \cdot d)$ &  33x & 30.8 & 5.1 \\ \hline
\multirow{1}{*}{NPC (ours)} & 3-layer Masked Conv. & \multirow{1}{*}{Local} & \multirow{1}{*}{$\mathcal{O}(k \cdot  d^2)$} & 1x & 27.9  & 6.1 \\
\bottomrule
\end{tabular}
\begin{tablenotes}
\item{\textdagger} \small{\text{Frame-wised time complexity. $T$ denotes the sequence length, $d$ the representation dimension, and $k$ the kernel size. }}
\item{\textsection} \small{\text{Averaged time cost over 10K runs on a single GPU with PyTorch~\cite{paszke2019pytorch} without further optimization on all networks}}

\end{tablenotes}
\label{table:compare}
\end{threeparttable}
\vspace{-4pt}
\end{table*}

\vspace{-5pt}
\subsection{Comparison with other self-supervised representation }
\label{exp:compare}

In Table~\ref{table:compare}, we compare NPC with prior speech representation learning models, including CPC~\cite{oord2018representation}, APC family~\cite{chung2019unsupervised,chung2020improved,chung2020vector}, and MLM family~\cite{wang2020unsupervised,liu2020mockingjay} as introduced in Sec.~\ref{sec:intro}.
We note that utterance-wise zero mean unit variance normalization on log Mel spectrograms is more suitable for NPC (and potentially all other methods), but we use speaker-wise normalization following the previous work specifically in Table~\ref{table:compare} for a fair comparison to the reported results in~\cite{chung2020vector}.

\vspace{3pt}
\noindent \textbf{Efficiency.}
To study the speed advantage of NPC brought by the non-autoregressive and local-only dependent property, we first compare the time complexity and empirical inference speed to others as shown in Table~\ref{table:compare}.
For time complexity, we consider the worst-case complexity per frame in terms of the input sequence length $T$, the representation dimension $d$, and the convolution kernel size $k$ for the NPC model.\footnote{We treat the depth of models $c$ as a constant since all models discussed in this paper have $c \ll T$ and $c \ll d$.}
For empirical inference time, we consider the averaged running time over 10K runs for all models with fixed sequence length $T=1000$ (approximately corresponded to a 10-second utterance), $d=512$, and a batch size of 32.

For NPC, the time complexity is $\mathcal{O}(k \cdot  d^2)$ since representation at any time step has a fixed-size receptive field depending on $k$, which is independent of the sequence length $T$.
We set the average running time of a 3-layer NPC as the standard (denoted "1x" in Table~\ref{table:compare}) and compare it against other methods.
For APC and CPC based methods, the worst case is the representation at the end of the sequence which must process through all $T$ inputs, resulting in the complexity $\mathcal{O}(T \cdot  d^2)$.
With the choice of 3-layer GRU, we observed 29 times longer inference time on APC and CPC models.
For RNN-MLM, the time complexity is identical to the previous case since the representation is the combination of 2 GRU hidden states.
However, in practice, bi-directional autoregressive representations can be up to 72 times slower than NPC without further optimization.
For the Transformer-MLM, the time complexity is $\mathcal{O}(T^2 \cdot  d)$ since each representation is a weighted sum of the complete sequence of hidden states of transformer encoders as noted in~\cite{vaswani2017attention}.
As speech signals are generally longer ($T>d$), we observed a slightly longer inference time than APC/CPC models.\footnote{In practice, this can be addressed by downsampling the feature sequence at the cost of making frame-wised representation unavailable.}

\vspace{3pt}
\noindent \textbf{Effectiveness.}
Given that NPC provides a significantly faster inference, we now take a look into the accessibility of speaker characteristics and phonetic information comparing to others.
In the task of speaker classification, representation from NPC produced a 6.1\% error rate where the best from Transformer-MLM is 1\% better.
This suggested that NPC may not be as effective as other representation models when the task explicitly requires global information.
For phone classification, which depends less on global information comparing to speaker classification, we observed a better performance compared to other methods, indicating that NPC can be applied for tasks focusing on local dependency without a trade-off.

However, we note that NPC is \textit{not} the best in terms of accessibility as a higher speaker error rate is observed.
In addition, we find that a lower PER of 25.6\% can be obtained from VQ-APC (v.s. 27.2\% from NPC in Table~\ref{table:ablation}) when the surface feature is utterance-wised normalized.
Nevertheless, the fast NPC provides a better opportunity for adapting large scale training and application in different downstream tasks without sacrificing much of performance.

\vspace{-6pt}
\subsection{Analysis on NPC}

Conceptually, NPC relied on local context information to predict the target frame without seeing itself.
This idea of learning contextual embedding based on the local neighbors in the sequence have been found useful in the field of learning word embedding~\cite{mikolov2013efficient,devlin2019bert} and have been extended to speech representation learning before~\cite{milde2018unspeech,peters2018deep,ling2020deep,chuang2020speechbert,song2020speech}.
However, we highlight that NPC has masking defined explicitly in the model and adopts simple reconstruction loss, making it different from other speech representation learning methods.

\begin{figure}[t]
\centerline{\includegraphics[width=7.5cm]{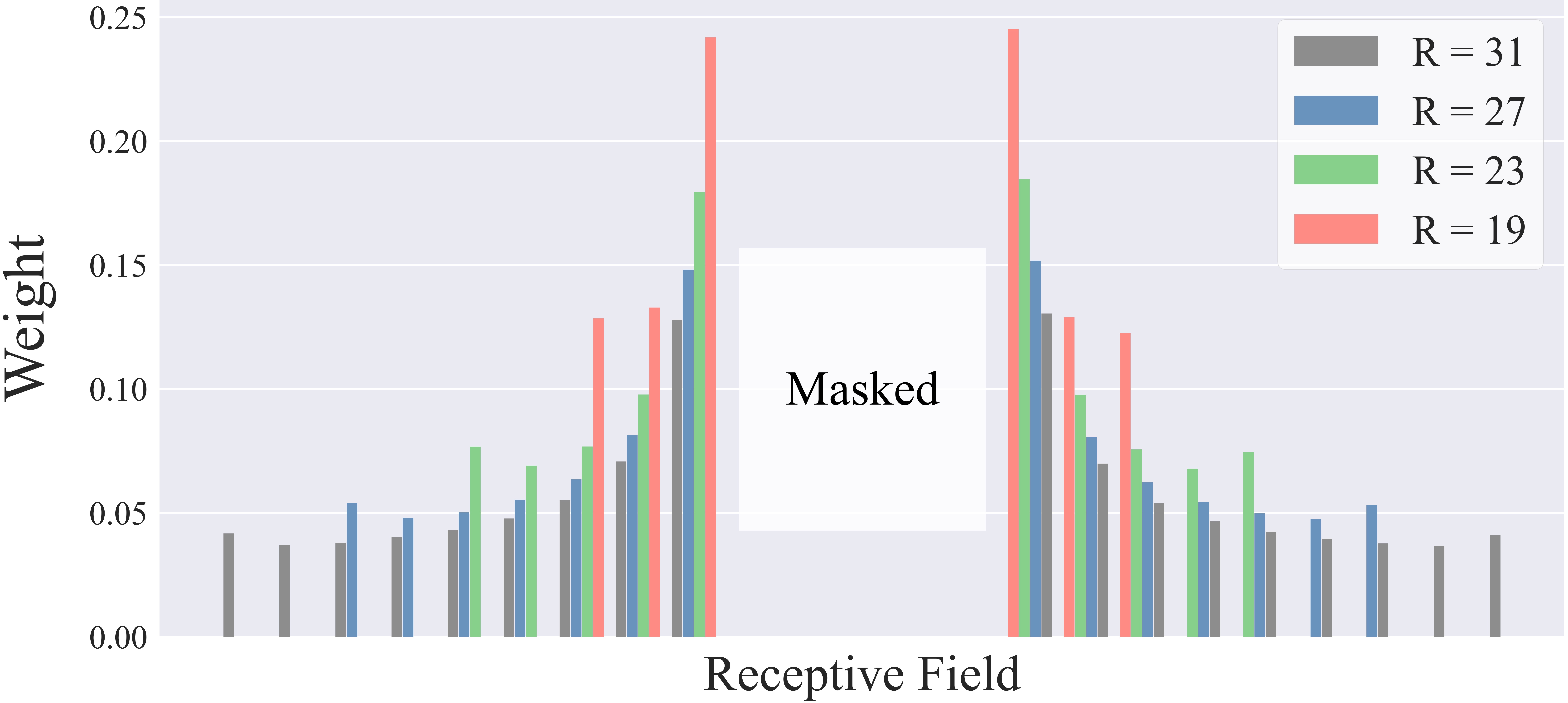}}
\vspace{-7pt}
\caption{\small Normalized magnitude of weights of CNN in Masked ConvBlock with different receptive field size $R$.}
\vspace{-15pt}
\label{fig:weight}
\end{figure}

To better understand how NPC derives representation from speech, we take the Masked ConvBlock kernels from the pre-trained 2-layer NPC model of different receptive fields $R$ and compute the magnitude of these kernel weights at the second layer.
This can be view as the importance of the adjacent frames of the target learned by NPC for generating speech representation.
Results are normalized and visualized in Fig.~\ref{fig:weight}.

Unsurprisingly, frames right next to the masked input always possess the largest part of total magnitude in kernel weights, indicating they are always the most important part for NPC to produce representation.
In addition, the inputs farthest from the target usually have less than 5\% of the total magnitude.
This further supported our point of view that local dependency is sufficient for learning effective speech representation.

\vspace{-8pt}
\section{Conclusion}
\label{sec:conclusion}
\vspace{-4pt}

In this work we pointed out the autoregressive and globally dependent property of different self-supervised methods that caused run time bottleneck.
With a simple objective, the proposed Non-Autoregressive Predictive Coding (NPC) can significantly speed up the inference time required for speech representation.
This is done by learning only from the local dependency of speech with a fix-sized receptive field.
Additionally, target-related information restriction is necessary and implemented by the proposed Masked ConvBlock.
In our experiments, we examined and discussed the importance of each design of NPC to demonstrate the robustness of the proposed framework.
Moreover, evaluation on representation learned and analysis on the model trained were carried out to support the conclusion that speech representation can be obtained more efficiently without hurting downstream tasks by NPC.

\newpage
\vfill\pagebreak
\bibliographystyle{IEEEbib}
{\bibliography{strings,refs}}

\begin{thebibliography}{10}

\bibitem{oord2018representation}
Aaron van~den Oord, Yazhe Li, and Oriol Vinyals,
\newblock ``Representation learning with contrastive predictive coding,''
\newblock {\em arXiv preprint arXiv:1807.03748}, 2018.

\bibitem{riviere2020unsupervised}
Morgane Rivi{\`e}re, Armand Joulin, Pierre-Emmanuel Mazar{\'e}, and Emmanuel
  Dupoux,
\newblock ``Unsupervised pretraining transfers well across languages,''
\newblock in {\em ICASSP}, 2020.

\bibitem{schneider2019wav2vec}
Steffen Schneider, Alexei Baevski, Ronan Collobert, and Michael Auli,
\newblock ``wav2vec: Unsupervised pre-training for speech recognition,''
\newblock in {\em Interspeech}, 2019.

\bibitem{baevski2020vq}
Alexei Baevski, Steffen Schneider, and Michael Auli,
\newblock ``{vq-wav2vec}: Self-supervised learning of discrete speech
  representations,''
\newblock in {\em ICLR}, 2020.

\bibitem{baevski2020wav2vec}
Alexei Baevski, Henry Zhou, Abdelrahman Mohamed, and Michael Auli,
\newblock ``wav2vec 2.0: A framework for self-supervised learning of speech
  representations,''
\newblock in {\em NeurIPS}, 2020.

\bibitem{chung2019unsupervised}
Yu-An Chung, Wei-Ning Hsu, Hao Tang, and James Glass,
\newblock ``An unsupervised autoregressive model for speech representation
  learning,''
\newblock in {\em Interspeech}, 2019.

\bibitem{pascual2019learning}
Santiago Pascual, Mirco Ravanelli, Joan Serr{\`a}, Antonio Bonafonte, and
  Yoshua Bengio,
\newblock ``Learning problem-agnostic speech representations from multiple
  self-supervised tasks,''
\newblock in {\em Interspeech}, 2019.

\bibitem{wang2020unsupervised}
Weiran Wang, Qingming Tang, and Karen Livescu,
\newblock ``Unsupervised pre-training of bidirectional speech encoders via
  masked reconstruction,''
\newblock in {\em ICASSP}, 2020.

\bibitem{liu2020mockingjay}
Andy Liu, Shu-Wen Yang, Po-Han Chi, Po-Chun Hsu, and Hung-Yi Lee,
\newblock ``Mockingjay: Unsupervised speech representation learning with deep
  bidirectional {Transformer} encoders,''
\newblock in {\em ICASSP}, 2020.

\bibitem{chung2020improved}
Yu-An Chung and James Glass,
\newblock ``Improved speech representations with multi-target autoregressive
  predictive coding,''
\newblock in {\em ACL}, 2020.

\bibitem{chung2020vector}
Yu-An Chung, Hao Tang, and James Glass,
\newblock ``Vector-quantized autoregressive predictive coding,''
\newblock in {\em Interspeech}, 2020.

\bibitem{devlin2019bert}
Jacob Devlin, Ming-Wei Chang, Kenton Lee, and Kristina Toutanova,
\newblock ``{BERT}: Pre-training of deep bidirectional {Transformers}for
  language understanding,''
\newblock in {\em NAACL-HLT}, 2019.

\bibitem{vaswani2017attention}
Ashish Vaswani, Noam Shazeer, Niki Parmar, Jakob Uszkoreit, Llion Jones, Aidan
  Gomez, {\L}ukasz Kaiser, and Illia Polosukhin,
\newblock ``Attention is all you need,''
\newblock in {\em NIPS}, 2017.

\bibitem{chorowski2019unsupervised}
Jan Chorowski, Ron Weiss, Samy Bengio, and A{\"a}ron van~den Oord,
\newblock ``Unsupervised speech representation learning using wavenet
  autoencoders,''
\newblock {\em IEEE/ACM Transactions on Audio, Speech, and Language
  Processing}, vol. 27, no. 12, pp. 2041--2053, 2019.

\bibitem{van2017neural}
Aaron van~den Oord, Oriol Vinyals, et~al.,
\newblock ``Neural discrete representation learning,''
\newblock in {\em NIPS}, 2017.

\bibitem{panayotov2015librispeech}
Vassil Panayotov, Guoguo Chen, Daniel Povey, and Sanjeev Khudanpur,
\newblock ``{LibriSpeech}: An {ASR} corpus based on public domain audio
  books,''
\newblock in {\em ICASSP}, 2015.

\bibitem{kingma2015adam}
Diederik Kingma and Jimmy Ba,
\newblock ``Adam: A method for stochastic optimization,''
\newblock in {\em ICLR}, 2015.

\bibitem{paul1992design}
Douglas Paul and Janet Baker,
\newblock ``The design for the {Wall Street Journal}-based {CSR} corpus,''
\newblock in {\em Speech and Natural Language Workshop}, 1992.

\bibitem{paszke2019pytorch}
Adam Paszke, Sam Gross, Francisco Massa, Adam Lerer, James Bradbury, et~al.,
\newblock ``{PyTorch}: An imperative style, high-performance deep learning
  library,''
\newblock in {\em NeurIPS}, 2019.

\bibitem{mikolov2013efficient}
Tomas Mikolov, Kai Chen, Greg Corrado, and Jeffrey Dean,
\newblock ``Efficient estimation of word representations in vector space,''
\newblock {\em arXiv preprint arXiv:1301.3781}, 2013.

\bibitem{milde2018unspeech}
Benjamin Milde and Chris Biemann,
\newblock ``Unspeech: Unsupervised speech context embeddings,''
\newblock in {\em Interspeech}, 2018.

\bibitem{peters2018deep}
Matthew Peters, Mark Neumann, Mohit Iyyer, Matt Gardner, Christopher Clark,
  Kenton Lee, and Luke Zettlemoyer,
\newblock ``Deep contextualized word representations,''
\newblock in {\em NAACL-HLT}, 2018.

\bibitem{ling2020deep}
Shaoshi Ling, Yuzong Liu, Julian Salazar, and Katrin Kirchhoff,
\newblock ``Deep contextualized acoustic representations for semi-supervised
  speech recognition,''
\newblock in {\em ICASSP}, 2020.

\bibitem{chuang2020speechbert}
Yung-Sung Chuang, Chi-Liang Liu, and Hung-Yi Lee,
\newblock ``{SpeechBERT}: Cross-modal pre-trained language model for end-to-end
  spoken question answering,''
\newblock in {\em Interspeech}, 2020.

\bibitem{song2020speech}
Xingchen Song, Guangsen Wang, Zhiyong Wu, Yiheng Huang, Dan Su, Dong Yu, and
  Helen Meng,
\newblock ``{Speech-XLNet}: Unsupervised acoustic model pretraining for
  self-attention networks,''
\newblock in {\em Interspeech}, 2020.

\end{thebibliography}
\end{document}